\documentclass{article}
\usepackage[T1]{fontenc} 
\usepackage[utf8]{inputenc} 
\usepackage{ismir,amsmath,cite,url}
\usepackage{amssymb}
\usepackage{graphicx}
\usepackage{color}
\def\lbd{}	

\usepackage{lineno}
\nolinenumbers

\title{Beat-Aligned Spectrogram-to-Sequence Generation of Rhythm-Game Charts}

\threeauthors
  {Jayeon Yi} {University of Michigan \\ {\tt jayeonyi@umich.edu}}
  {Sungho Lee} {Seoul National University \\ {\tt sh-lee@snu.ac.kr}}
  {Kyogu Lee}{Seoul National University \\ {\tt kglee@snu.ac.kr} }

\def\authorname{J. Yi, S. Lee, and K. Lee}

\begin{document}

\maketitle
\begin{abstract}
In the heart of "rhythm games" - games where players must perform actions in sync with a piece of music - are "charts", the directives to be given to players. We newly formulate chart generation as a sequence generation task and train a Transformer using a large dataset. We also introduce tempo-informed preprocessing and training procedures, some of which are suggested to be integral for a successful training. Our model is found to outperform the baselines on a large dataset, and is also found to benefit from pretraining and finetuning.


\end{abstract}
\section{Introduction}\label{sec:introduction}


\emph{Rhythm games} - games where players perform specific actions in sync with a piece of music - are enjoyed by millions of players worldwide \cite{ddc}. A number of past researches have investigated the automated design of the sequence of actions, or \emph{charts}, given the music \cite{ddc, halina, genelive}. As the ground truth depends on the desired ``difficulty'' of the chart as well as the music, the past approaches fed spectrograms and a encoded difficulty parameter to deep neural networks (DNNs), inferring if each ``audio frame'' contains a chart event. However, such approaches were observed to exhibit binary class imbalance due to the temporal sparsity of ground-truth chart events, especially for low-difficulty charts \cite{ddc}. As a remedy Takada \emph{et al.} \cite{genelive} successfully proposed to incorporate tempo information, but the sparsity itself persists. 

We newly formulate chart generation as a conditional sequence generation task, thus removing the binary class imbalance. We beat-align and normalize the length of our four-beat training samples, train a Transformer \cite{transformer} on our large dataset, and fine-tune to the common benchmarks. We also examine how our approach scales with more data, and how our beat-aligned training procedures may be integral to a successful training. Our method outperforms previous approaches in terms of micro-F1 scores. We share demos, code, and pretrained models on our companion website \footnote{ \url{https://stet-stet.github.io/goct} }.



\section{Tempo-Informed Preprocessing}


\begin{figure}
 \centerline{
 \includegraphics[width=\columnwidth]{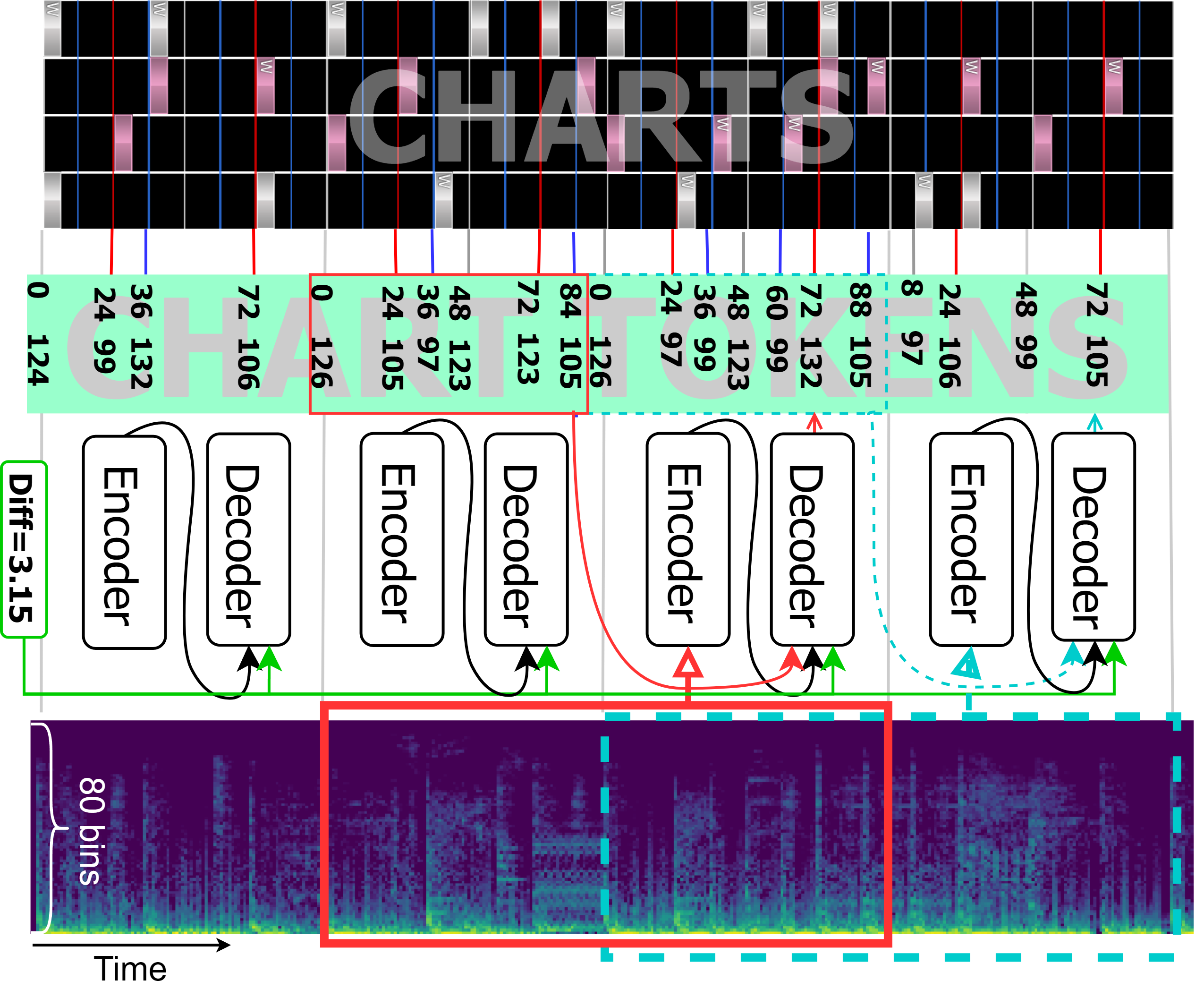}}
 \caption{A schematic of our approach. Four beats of beat-aligned log-Mel spectrograms, a difficulty value, and \emph{chart tokens} for the two former beats are used to infer chart tokens for the two latter beats.}
 \label{fig:pipeline}
\end{figure}



\subsection{Dataset}


To facilitate training, we build a large dataset on the popular ``four-keys''(4k) mode of ``osu!mania'', a game where players follow on-screen instructions to tap, hold, or stop holding each of the four "keys" in sync with a piece of accompanying music. We collected a total of 14648 charts aligned with a total of 3166 stereo-audio songs, the alignment of which were peer-reviewed by seasoned community creators. 6781 charts and 2004 songs remained after only leaving ``4k'' charts without off-beat tempo changes or extremely dense (>25 events/beat) parts. We split the dataset into three splits as represented in Table 1, and analyze the frequency of various beat groups in Table 3. 

The Fraxtil and ITG datasets \cite{ddc} were similarly preprocessed for benchmarking purposes. These datasets are based on a game with a syntactically same set of controls as ``osu!mania 4k''. However, while the former provides algorithmic chart difficulty assessments as a floating-point number, 
the latter provides human-labeled integers.

\subsection{Tempo-Informed Preprocessing}

Our new preprocessing procedures involve the beat-aligning of training samples - a classic idea in MIR literature (e.g. \cite{ellis2012large, dieleman2011audio}). We processed all music into mono-channel log-Mel spectrograms with 80 Mel bins, FFT and window sizes of 512, and hop sizes corresponding to 1/48 of a beat at any given moment. Thus, a piece of music with a length of 400 beats, regardless of tempo, would be processed into a matrix $S \in \mathbb{R}^{19200\times80}$. 

\subsection{Tokenization}

We use \emph{chart tokens} to formulate chart generation as a sequence generation problem. An eight-beat example chart is translated into chart tokens on the top of Fig. \ref{fig:pipeline}. Each chart event is expressed in two tokens - a \emph{time token}  (0-95) denoting its temporal position, and an \emph{action token} (97-176) identifying which of the 80 acceptable player actions are associated with the event. A ``separator'' token can be used to set a reference for time tokens. 
Finally, The end-of-sequence (EOS) token is used to pad the sequence when needed. 
Chart tokens resemble MIDI tokens in symbolic music literature \cite{MT3, musictransformer}, but are much simpler due to the narrower variety of chart events compared to MIDI.

\section{Generating charts} 

\begin{table}
 \begin{center}
 \begin{tabular}{l|rrr}
  \hline
  \hline
  Split & \textbf{Train} & \textbf{Valid} & \textbf{Test} \\
  \hline
  \# songs  & 1598(73 hrs) & 199(9.3 hrs) & 207(9.8 hrs) \\
  \# charts & 5494(219 hrs) & 634(26 hrs) & 643(25 hrs) \\
  \# beats & 2.24M & 252k & 255k \\
  \hline
  \hline
 \end{tabular}
\end{center}
\vspace{-3mm}
 \caption{A summary of our "osu!mania" 4-key dataset. "\# beats" are equivalent to the number of samples in each set, i.e. 2.24M samples were used for training. This dataset is approximately 10x bigger than Fraxtil and ITG datasets.}
 \label{tab:dataset}
\end{table}


\begin{table}
 \begin{center}
 \begin{tabular}{l|ccc}
  \hline
  \hline
  Method & Fraxtil (\%) & ITG (\%) & osu! (\%)\\
  \hline
  CNN \cite{ddc} & 75.0 & 71.9 & 83.0 \\
  C-LSTM \cite{ddc} & 75.6 & 72.1 & 83.9 \\
  \hline
  Ours & 70.5 & 68.7 & \textbf{84.6} \\
  Ours (pretrained)  & \textbf{78.9} & \textbf{74.3} & -\\ 
  \hline
  \hline
 \end{tabular}
\end{center}
\vspace{-3mm}
 \caption{$F_1$ scores on Fraxtil, ITG, and osu!mania datasets. Pretraining was done on the ``osu!mania'' dataset.}
 \label{tab:compare}
\end{table}



\subsection{Model Architecture and Pipeline} 

We use encoder-decoder Transformers, which have enjoyed considerable success in musical sequence generation tasks \cite{MT3, musictransformer}. Our pipeline is depicted in Fig. \ref{fig:pipeline}. Four beats of log-Mel spectrograms are fed to the encoder, while the last seven chart tokens from the former two beats are concatenated with 48-dim difficulty value embeddings and fed to the decoder. Chart tokens for the latter two beats are autoregressively generated, and then similarly truncated and fed into the model again to produce tokens for the two beats to follow after. This continues until the end of the music is reached. We use three layers and an input dimension of 256, since larger models did not yield any increase in performance.

\subsection{Training}

For the proposed model, all training and validation samples are aligned to beat boundaries. We trained on each dataset for 10 epochs with the adam \cite{adam} optimizer with lr=2e-4, and bs=32. We used the cross-entropy loss with a label-smoothing of 0.02. We then took this model, ``pretrained'' on the ``osu!mania'' dataset, and then finetuned on the Fraxtil and ITG datasets for 4 epochs, with lr=2e-5. 

Models compared to the baselines were trained \emph{without action tokens}; the baseline models only infer temporal positions of events. All others are trained with action tokens.

\subsection{Evaluation, Metrics and Analysis}

\begin{table} 
 \begin{center}
 \begin{tabular}{l|ccccc}
  \hline
  \hline
  Beat Groups & 8th & 16th & 12th & 32nd & 24th \\
  \hline
  \% freq  & 31.3 & 20.7 & 3.1 & 1.8 & 1.3 \\
  \hline
  $F_1$ unaligned (\%) & 70.0 & 47.6 & 9.2 & 3.27 & 3.94 \\
  $F_1$ aligned (\%)  &  \textbf{87.9} & \textbf{64.0} & \textbf{59.2} & \textbf{21.5} & \textbf{26.3} \\
  \hline
  \hline 
 \end{tabular}
\end{center}
\vspace{-3mm}
 \caption{Micro-$F_1$ scores of aligned and unaligned models on ``osu!mania'' dataset for each timing group. Frequency of occurrence in dataset is denoted on the second row. }
 \label{tab:example}
\end{table}



Micro-$F_1$ scores, calculated from temporal locations of chart events, are presented on Table 2. While our approach falls short of the baselines on small datasets such as Fraxtil and ITG, it outperforms the baselines when trained on a large dataset such as ``osu!mania''. Moreover, pretraining on a large dataset significantly improves the scores. The results are reminiscent of past works which showed that large-scale training with a Transformer can trump inductive bias \cite{vit}. We conjecture that utilizing more data would help to further improve our approach.

Our approach is not without limitations - even the finetuned model fails catastrophically on generating low-difficulty charts of Fraxtil and ITG datasets. This is conjectured to be due to how ``osu!mania'' dataset rarely includes \emph{very} sparse charts like these. We conjecture that training on a more diverse set of data could be a solution.

\subsection{Ablation: Unaligned Training}

We train a model on the "osu!mania" dataset using non-beat-aligned four-beat-long training samples, keeping all other conditions the same. Table 3 shows micro-$F_1$ scores of both models for moderately-used time-tokens, grouped by their temporal offset relative to beat boundaries. Similar to \cite{ddc}, any time-token from the unaligned model was deemed correct if within 30ms of any ground-truth time-token. Table 3 shows that the aligned model exhibits stronger generalization. Moreover, It was qualitatively observed that the unaligned model would output temporally shifted outputs, with the shift changing over time. This result suggests that the beat alignment is vital to the successful training of our transformer-based models.

\bibliography{ISMIRtemplate}

\begin{thebibliography}{10}
\providecommand{\url}[1]{#1}
\csname url@samestyle\endcsname
\providecommand{\newblock}{\relax}
\providecommand{\bibinfo}[2]{#2}
\providecommand{\BIBentrySTDinterwordspacing}{\spaceskip=0pt\relax}
\providecommand{\BIBentryALTinterwordstretchfactor}{4}
\providecommand{\BIBentryALTinterwordspacing}{\spaceskip=\fontdimen2\font plus
\BIBentryALTinterwordstretchfactor\fontdimen3\font minus \fontdimen4\font\relax}
\providecommand{\BIBforeignlanguage}[2]{{%
\expandafter\ifx\csname l@#1\endcsname\relax
\typeout{** WARNING: IEEEtran.bst: No hyphenation pattern has been}%
\typeout{** loaded for the language `#1'. Using the pattern for}%
\typeout{** the default language instead.}%
\else
\language=\csname l@#1\endcsname
\fi
#2}}
\providecommand{\BIBdecl}{\relax}
\BIBdecl

\bibitem{ddc}
C.~Donahue, Z.~C. Lipton, and J.~McAuley, ``Dance dance convolution,'' in \emph{Proc. of the 34th International Conference on Machine Learning}, Sydney, Australia, 2017, pp. 1039--1048.

\bibitem{halina}
E.~Halina and M.~Guzdial, ``Taikonation: Patterning-focused chart generation for rhythm action games,'' in \emph{Proc. of the 16th International Conference on the Foundations of Digital Games}, 2021, pp. 1--10.

\bibitem{genelive}
A.~Takada, D.~Yamazaki, Y.~Yoshida, N.~Ganbat, T.~Shimotomai, N.~Hamada, L.~Liu, T.~Yamamoto, and D.~Sakurai, ``Gen{\'e}live! generating rhythm actions in love live!'' in \emph{Proceedings of the AAAI Conference on Artificial Intelligence}, vol.~37, no.~4, 2023, pp. 5266--5275.

\bibitem{transformer}
A.~Vaswani, N.~Shazeer, N.~Parmar, J.~Uszkoreit, L.~Jones, A.~N. Gomez, {\L}.~Kaiser, and I.~Polosukhin, ``Attention is all you need,'' \emph{Advances in neural information processing systems}, vol.~30, 2017.

\bibitem{ellis2012large}
D.~P. Ellis and B.-M. Thierry, ``Large-scale cover song recognition using the 2d fourier transform magnitude,'' 2012.

\bibitem{dieleman2011audio}
S.~Dieleman, P.~Brakel, and B.~Schrauwen, ``Audio-based music classification with a pretrained convolutional network,'' in \emph{Proceedings of the 12th International Society for Music Information Retrieval Conference (ISMIR-2011)}.\hskip 1em plus 0.5em minus 0.4em\relax University of Miami, 2011, pp. 669--674.

\bibitem{MT3}
J.~Gardner, I.~Simon, E.~Manilow, C.~Hawthorne, and J.~Engel, ``Mt3: Multi-task multitrack music transcription,'' in \emph{Proc. of the 10th International Conference on Learning Representations}, 2022.

\bibitem{musictransformer}
C.-Z.~A. Huang, A.~Vaswani, J.~Uszkoreit, I.~Simon, C.~Hawthorne, N.~Shazeer, A.~M. Dai, M.~D. Hoffman, M.~Dinculescu, and D.~Eck, ``Music transformer: Generating music with long-term structure,'' in \emph{Proceedings of the 6th International Conference on Learning Representations}, 2018.

\bibitem{adam}
D.~P. Kingma and J.~Ba, ``Adam: A method for stochastic optimization,'' \emph{arXiv preprint arXiv:1412.6980}, 2014.

\bibitem{vit}
A.~Dosovitskiy, L.~Beyer, A.~Kolesnikov, D.~Weissenborn, X.~Zhai, T.~Unterthiner, M.~Dehghani, M.~Minderer, G.~Heigold, S.~Gelly \emph{et~al.}, ``An image is worth 16x16 words: Transformers for image recognition at scale,'' \emph{arXiv preprint arXiv:2010.11929}, 2020.

\end{thebibliography}

\end{document}